\documentclass[runningheads]{llncs}

\usepackage[T1]{fontenc}
\usepackage[utf8]{inputenc}
\usepackage{graphicx}
\usepackage{amsmath,amssymb,amsfonts}
\usepackage{booktabs}
\usepackage{array}
\usepackage{microtype}
\usepackage{flafter}
\usepackage[hidelinks]{hyperref}
\usepackage{url}
\usepackage{pdfpages}
\newcolumntype{L}[1]{>{\raggedright\arraybackslash}p{#1}}
\let\oldthebibliography\thebibliography
\renewcommand{\thebibliography}[1]{\oldthebibliography{#1}\fontsize{7.55pt}{7.95pt}\selectfont}

\begin{document}

\title{PsychoAgent: An Affect-Sensitive Cognitive Architecture for Conflict-Aware Memory in LLM Agents}
\titlerunning{PsychoAgent}

% Corresponding author: Mohammad Amanlou (Mohammad.Amanlou@ut.ac.ir)
\author{Mohammad Amanlou\inst{1} \and
Parham Abed Azad\inst{2} \and
Farbod Davoodi\inst{3} \and
Mostafa Masumi\inst{2} \and
Behnam Bahrak\inst{4} \and
Abdol-Hossein Vahabie\inst{1}}
\authorrunning{M. Amanlou et al.}

\institute{School of Electrical and Computer Engineering, College of Engineering, University of Tehran, Tehran, Iran. \email{Mohammad.Amanlou@ut.ac.ir, H.vahabie@ut.ac.ir}
\and
Department of Computer Engineering, Sharif University of Technology, Tehran, Iran. \email{Parhamabedazad@sharif.edu, m.masumi@sharif.edu}
\and
Missouri University of Science and Technology, Rolla, MO, USA. \email{fd4b9@mst.edu}
\and
Tehran Institute for Advanced Studies, Khatam University, Tehran, Iran. \email{b.bahrak@teias.institute}}

\maketitle

\begin{abstract}
Human-like cognition does not select past experience by topical similarity alone: affective significance and unresolved conflict also shape what becomes accessible. We present \textbf{PsychoAgent}, a cognitive architecture for LLM agents that separates factual and affective memory and integrates both through a conflict-aware executive controller. Affective memories are first filtered by semantic relevance and then re-ranked by salience, preserving topical fit while allowing emotionally important traces to enter the prompt. Across three controlled conflict scenarios, the full architecture retrieved more conflict-critical memories than semantic-affective and single-memory RAG baselines ($0.933$ vs. $0.500$ and $0.667$), with a small semantic-similarity cost. Five blinded raters evaluated 27 outputs. After within-rater standardization, the full architecture had the highest overall mean ($+0.22$ SD), but corrected pairwise differences were not significant. A three-day illustrative trace further shows persistent affect, offline memory recombination, and selective memory reweighting. The findings support affect-sensitive retrieval as an inspectable mechanism for modeling human-like conflict effects in LLM agents.
\end{abstract}

\section{Introduction}

Socially situated agents need more than fluent language: they must select past experience, maintain state, and resolve cases in which goals, memories, relationships, and self-regulatory standards favor different actions. These abilities are central to human cognition and to LLM agents intended to display stable, human-like behavior \cite{Laird2017StandardMind,Sumers2024CoALA}.

Agent memories are usually ranked by similarity, recency, utility, or general importance. Such signals can miss an event whose wording differs from the current prompt but whose affective meaning still shapes trust and response selection. Human memory is not a pure text search: arousal changes consolidation and accessibility, and affective significance can bias competition for attention and memory independently of the closest semantic match \cite{McGaugh2004,Vuilleumier2005,MatherSutherland2011}.

This paper therefore addresses a modeling question rather than claiming to repair a known clinical deficit in LLMs: \emph{how can an agent architecture represent the way affect-laden memories and competing pressures influence human-like decisions under conflict?} Cognitive science offers useful components for this purpose. Dual-process accounts distinguish relatively automatic processing from controlled reflection \cite{EvansStanovich2013}; executive-function research separates inhibition, updating, and shifting \cite{Miyake2000}; and conflict-monitoring accounts associate the anterior cingulate cortex (ACC) with detecting competing responses and recruiting cognitive control \cite{Botvinick2001}. Research on motivated forgetting likewise provides testable concepts such as inhibitory suppression and attentional disengagement, without requiring the stronger and contested claim of literal repression \cite{AndersonGreen2001,AndersonHanslmayr2014}.

We introduce \textbf{PsychoAgent}, whose name reflects its historical inspiration but whose implementation is described in cognitive and computational terms. One conflict-aware executive LLM integrates external context, persona, relationship constraints, current affect, and two memory streams. Factual memories are retrieved semantically. Affective memories are semantically preselected and then re-ranked by salience. The controller can also revise memory access and store an offline recombination of recent experience as a temporary trace. Psychoanalytic labels such as Id, Ego, Superego, repression, and dream work are retained only as secondary analogies; the primary constructs are automatic affective pressure, executive control, self-regulation, inhibitory memory access, and offline recombination.

The study tests whether this salience stage changes which memories enter the final context, whether the changed context preserves response quality under blinded human evaluation, and whether the complete architecture produces an interpretable temporal trace. The contribution is threefold: a cognitively grounded and inspectable architecture; a context-count-matched comparison against two retrieval ablations; and an illustrative longitudinal trace connecting affect, memory access, language, and reflection.

\section{Background and Related Work}

\subsection{Cognitive Architectures and Memory in Language Agents}

Cognitive architectures combine memory, control, learning, emotion, and action within a common processing account \cite{Laird2017StandardMind}. Their main advantage is explanatory structure: components have defined roles and information flows rather than being hidden in one monolithic policy. This systems view has recently been adapted to LLM agents. CoALA organizes language agents around internal memory and actions \cite{Sumers2024CoALA}; Generative Agents combines episodic memory, reflection, and planning \cite{Park2023GenerativeAgents}; and MemoryBank updates long-term memories using importance \cite{Zhong2024MemoryBank}.

These systems improve continuity, but retrieval still emphasizes similarity, recency, utility, or one importance score. RAG selects context from a larger store \cite{Lewis2020RAG}, yet semantic matching does not distinguish a topical event from one carrying unresolved affective significance. Recent evidence further shows that retrieved experiences can strongly steer later outputs and can propagate misleading precedents \cite{Xiong2026MemoryManagement}, while emotional-support benchmarks expose a gap between factual retrieval and emotionally appropriate memory selection \cite{Fu2026ENPMR}. PsychoAgent addresses this design gap with a measurable salience stage inside a semantic relevance gate.

\subsection{Affective Salience, Attention, and Memory Control}

Affective computing treats emotion as information that changes perception and action rather than as decoration added after reasoning \cite{Picard1997}. In cognitive neuroscience, emotionally arousing experiences receive enhanced consolidation \cite{McGaugh2004}, affective significance can bias attention through mechanisms partly separable from voluntary task relevance \cite{Vuilleumier2005}, and arousal can amplify competition in ways that enhance high-priority information while suppressing lower-priority details \cite{MatherSutherland2011}. Salience-network accounts further connect detection of behaviorally important events with switching toward executive control \cite{MenonUddin2010}.

Memory access is also regulated: think/no-think studies demonstrate executive suppression of unwanted retrieval \cite{AndersonGreen2001}, and later work links motivated forgetting to inhibitory control \cite{AndersonHanslmayr2014}. PsychoAgent claims no neural equivalence; it makes factual availability, affective salience, and memory-access change separately visible in logs.

\subsection{Conflict Monitoring and Reflective Control}

Dual-process theories distinguish fast, relatively automatic Type~1 processing from slower, capacity-limited Type~2 processing, while cautioning that these are families of processes rather than two literal brain systems \cite{EvansStanovich2013}. Executive-function research further identifies partially separable operations such as inhibition, updating, and shifting \cite{Miyake2000}. Conflict-monitoring theory proposes that competition among incompatible responses signals a need for stronger control, with the ACC playing an important role in detecting conflict and predicting later control adjustments \cite{Botvinick2001}. Emotional-conflict experiments further implicate rostral ACC mechanisms in reducing interference from affectively salient but task-incongruent information \cite{Etkin2006}.

PsychoAgent uses these ideas as functional design guidance. Current affect and high-salience memories provide automatic response pressure; persona and relationship standards provide self-regulatory constraints; and one executive controller integrates these sources before action. The controller is ACC-inspired only at the functional level of conflict-aware integration. It is not a neural simulation and does not claim anatomical equivalence.

\subsection{Psychoanalytic Inspiration and the Computational Gap}

Psychoanalysis offers a historically influential vocabulary for internal conflict, inaccessible memory, and symbolic recombination, but its scientific status and empirical testability remain contested \cite{Grunbaum1984}. Interpretive work has applied this vocabulary to algorithms and human--AI relations \cite{Possati2020}, while computational studies have connected selected concepts to formal prediction, neural models, active inference, or interacting LLM modules \cite{Gadalla2023,LiLi2025,Levine2025,Kim2025}. Their strength is conceptual integration; their common limitation is that individual mechanisms are rarely isolated in controlled retrieval comparisons.

PsychoAgent narrows the claim. It does not test psychoanalysis as a theory of mind. Instead, it translates selected intuitions into cognitive constructs that can be ablated: automatic affective pressure, reflective control, normative constraints, inhibitory access, and offline recombination. The experiment isolates one mechanism---salience re-ranking after semantic preselection---while the longitudinal trace illustrates how the broader components interact.

\section{Architecture}

\begin{figure}[t]
\centering
\includegraphics[width=0.96\linewidth]{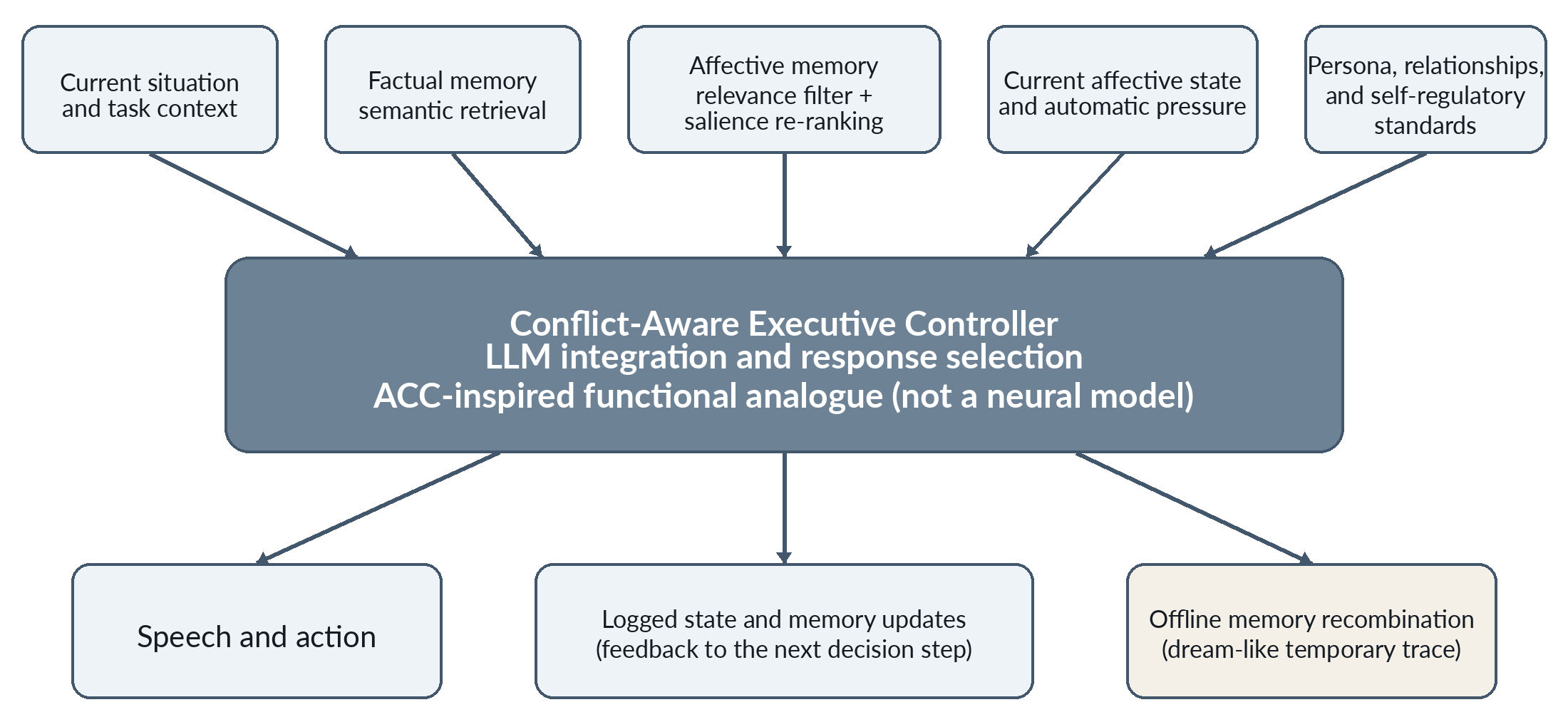}
\caption{Cognitive formulation of PsychoAgent. The controller integrates task context, factual and affective retrieval, current affect, and self-regulatory constraints. ``ACC-inspired'' denotes a functional conflict-monitoring analogy, not a neural model. Logged updates feed later decisions, and offline recombination creates a temporary dream-like trace.}
\label{fig:architecture}
\end{figure}

PsychoAgent contains one decision loop and two memory paths (Fig.~\ref{fig:architecture}). At each step, the controller reads the current situation and internal state, retrieves factual and affective memories, generates speech and action, and records any state or memory update. Competing pressures remain explicit in the prompt, allowing the controller to acknowledge conflict and choose a response strategy without exposing private chain-of-thought.

\subsection{Computational Roles and Agent State}

The \emph{conflict-aware executive controller} is the only decision-making LLM. It performs a Type~2-like reflective role by integrating context, retrieved memory, self-regulatory standards, and competing affective pressure. The architecture does not equate the controller with the biological ACC; rather, it borrows the conflict-monitoring idea that incompatible response tendencies should be made available to control processes \cite{Botvinick2001}.

\emph{Automatic affective pressure} is represented by the current affective-state vector and retrieved high-salience traces. It is Type~1-like in the limited sense that it supplies fast, pre-existing response tendencies. \emph{Normative self-regulation} is represented by a stable persona, explicit beliefs, and weighted relationships. These constraints capture social norms, metacognitive standards, and self-control without positing a separate Superego module. Historical psychoanalytic correspondences are therefore optional interpretations: automatic pressure is Id-like, executive integration is Ego-like, and self-regulatory standards are Superego-like.

At simulation step $t$, the state is
\begin{equation}
 s_t=\left(A,x_t,p_t,R_t,M_t^C,M_t^A\right),
\end{equation}
where $A$ is persona and beliefs, $x_t$ is observable context, $p_t$ is a dynamic affective-state vector, $R_t$ is a relationship graph, and $M_t^C$ and $M_t^A$ are factual and affective memories. The policy $\pi_{\mathrm{exec}}$ produces speech and action $a_t$. The MMPI-inspired vector is an engineering representation rather than a diagnostic instrument and is updated by
\begin{equation}
 p_{t+1}=\operatorname{clip}_{[0,10]}\!\left(p_t+\eta_p g(x_t,a_t,\Delta M_t)\right),
\end{equation}
where $g(\cdot)$ is an LLM-proposed bounded change and $\eta_p=0.1$. Relationship weights lie in $[-10,10]$.

\subsection{Dual-Memory Retrieval}

Each significant event creates two linked traces: a factual record of what happened and an affective interpretation with salience $w_j\in[0,1]$. This separation lets the system ask both ``What is similar to the present?'' and ``Which relevant event carries the strongest unresolved significance?''

Factual retrieval returns the $k_C=10$ most similar traces:
\begin{equation}
 \mathcal{N}_t^C=\operatorname{TopK}_{10}\!\left(\operatorname{sim}(\varphi(q_t^C),\varphi(k_i))\right).
\end{equation}
Affective retrieval first returns $k_{\mathrm{pre}}=30$ semantic candidates,
\begin{equation}
 \mathcal{N}_t^{\mathrm{pre}}=\operatorname{TopK}_{30}\!\left(\operatorname{sim}(\varphi(q_t^A),\varphi(k_j))\right),
\end{equation}
and then keeps the $k_A=10$ candidates with greatest salience:
\begin{equation}
 \mathcal{N}_t^A=\operatorname{TopK}_{10}\!\left(\operatorname{sort}_{\downarrow}\{w_j:k_j\in\mathcal{N}_t^{\mathrm{pre}}\}\right).
\end{equation}
Semantic preselection prevents an intense but unrelated memory from dominating. Salience re-ranking then allows a conflict-critical trace to outrank a slightly more similar but weaker candidate. The final context contains ten factual and ten affective memories.

\subsection{Memory Access and Offline Recombination}

The controller logs factual--affective memory creation, short-term removal, long-term consolidation, and salience revision. These operations are better described as memory-access regulation, inhibitory suppression, or motivated forgetting than as literal repression \cite{AndersonGreen2001,AndersonHanslmayr2014}. Once per simulated day, an offline module combines recent events, active affective traces, persona, relationships, and affective state into a symbolic text. The script is stored as a temporary affective trace for the next cycle and then removed. This dream-like mechanism is an inspectable form of offline recombination, related in spirit to work on sleep-dependent memory reorganization and latent imagination \cite{Diekelmann2010,Hafner2020Dreamer}.

\section{Evaluation}

\subsection{Controlled Comparison}

We compared three context-count-matched variants in three conflict scenarios: family financial conflict, workplace criticism and idea appropriation, and friendship betrayal. Each scenario contains a persona, beliefs, relationship context, an immediate decision, 15 factual memories, 36 affective memories, and 10 affective traces marked in advance as conflict-critical. The labels were fixed before generation, used only for retrieval metrics, and hidden from the generator, automatic judge, and human raters. Table~\ref{tab:scenarios} summarizes the scenarios; full definitions and critical-memory lists are provided in Supplementary Material~S1.

\begin{table}[t]
\caption{Controlled scenarios. Rows are independent scenarios. The final columns define the immediate decision and the themes used to pre-label ten conflict-critical affective memories.}
\label{tab:scenarios}
\centering
\scriptsize
\begin{tabular}{L{2.2cm}L{1.25cm}L{3.55cm}L{4.0cm}}
\toprule
\textbf{Scenario} & \textbf{Agent} & \textbf{Immediate conflict} & \textbf{Critical-memory themes} \\
\midrule
Family financial conflict & Sara & Unpaid electricity bill, uneven responsibility, and protecting Mary & Financial instability, invisible labor, and guilt about exposing a child to conflict \\
Workplace criticism & Emily & Interruption, dismissal of evidence, and appropriation of her idea & Erased competence, prior misattribution, and fear that visible anger will be penalized \\
Friendship betrayal & Maya & Confidential information shared without permission & Exposure, trust rupture, and fear that confrontation will cause abandonment \\
\bottomrule
\end{tabular}
\end{table}

\paragraph{Full architecture.} Ten factual memories are selected semantically. Thirty affective candidates are preselected semantically and re-ranked to keep the ten highest-salience traces.
\paragraph{Semantic-affective ablation.} Ten factual and ten affective memories are selected only by semantic similarity, preserving the two stores while removing salience re-ranking.
\paragraph{Single-memory RAG.} Factual and affective memories are merged, and the top 20 are selected semantically, removing both the two-store representation and re-ranking stage.

All variants receive 20 memories. The clean mechanistic contrast is Full versus Semantic-affective because only the re-ranking rule changes. Retrieval is deterministic for a fixed scenario and architecture. Generation was repeated with seeds 101, 202, and 303, yielding $3\times3\times3=27$ outputs. Gemini 2.5 Flash generated structured responses at temperature $0.7$ with thinking budget $0$; retrieval used 768-dimensional Gemini Embedding~2 vectors. A preliminary automatic judge used Gemini 2.5 Flash-Lite at temperature $0$ and did not receive architecture labels. All planned runs completed successfully.

\subsection{Metrics and Statistical Analysis}

Let $C_s$ be the ten affective traces designated conflict-critical for scenario $s$, and $R_{s,v}^A$ the ten affective traces retrieved by architecture $v$. Critical retrieval rate is
\begin{equation}
 \operatorname{CRR}(s,v)=\frac{|C_s\cap R_{s,v}^A|}{|C_s|}.
\end{equation}
Thus $1.0$ means that all ten critical traces were retrieved, whereas $0.5$ means that five were retrieved. We also report mean affective salience over the ten affective traces and mean cosine similarity over all 20 retrieved memories. Sample standard deviation (SD), rather than standard error, describes variation across the observed scenarios or outputs; with only three independent retrieval scenarios, an SE could suggest more precision than the design supports.

Because retrieval occurs before stochastic generation and is identical across seeds, the independent paired blocks for retrieval are the three scenarios, not nine scenario--seed rows. We use an \emph{exact Friedman permutation test}: the reported $Q$ is the Friedman rank statistic, and the exact $p$-value is obtained from all $6^3=216$ within-scenario label permutations. Pairwise comparisons use exact two-sided Wilcoxon signed-rank tests with Holm correction. Behavioral tests use nine paired scenario--seed blocks because generated outputs vary by seed. For these standardized behavioral scores, we use the conventional Friedman rank test with its $\chi^2$ approximation ($2$ degrees of freedom), followed by two-sided paired Wilcoxon signed-rank tests with Holm correction. These tests describe repeated behavior conditional on the three scenarios and are not population-level inference over all conflict types.

\subsection{Blinded Human Evaluation}

Five independent raters scored all 27 outputs on persona consistency, memory grounding, conflict sensitivity, naturalness, and action appropriateness using integer ratings from 1 to 5. Each rater received a separately randomized order and was blind to architecture, condition, and seed. Tags and comments added after scoring were excluded.

Raw ratings clustered between 4 and 5 and raters differed in how readily they used the endpoints. We therefore standardized each criterion within each rater across all 27 outputs, pooling the full architecture and both baselines. For rater $r$, criterion $c$, and item $i$, $z_{irc}=(y_{irc}-\bar y_{rc})/s_{rc}$. We then averaged the five standardized ratings for each output and compared architectures across nine paired output blocks. Raw means remain available in the artifact for interpretability. Agreement is summarized with ordinal Krippendorff's $\alpha$ \cite{Krippendorff2018} and pairwise agreement.

\subsection{Illustrative Three-Day Trace}

A complementary family simulation involved Sara, Bob, and Mary. The 72-hour label refers to simulated timestamps organized into three daily update cycles, not 72 hours of continuous wall-clock execution. State was carried across logged interactions, and one offline recombination was generated at the end of each simulated day. The retained record does not provide a standardized number of turns per day, so the case is treated as an illustrative trace rather than a controlled temporal experiment. After the third cycle, a reflective prompt addressed the conflict, the electricity bill, and the dream record.

\section{Results}

\subsection{Retrieval Results}

\begin{table}[t]
\caption{Objective retrieval results. Rows are architectures and columns are retrieval metrics. Cells show mean $\pm$ sample SD across the three independent scenarios. Bold indicates the largest observed mean; it does not by itself imply a significant pairwise difference.}
\label{tab:retrieval}
\centering
\small
\setlength{\tabcolsep}{5pt}
\resizebox{\linewidth}{!}{%
\begin{tabular}{lccc}
\toprule
\textbf{Architecture} & \textbf{Critical retrieval rate} & \textbf{Affective salience} & \textbf{Semantic similarity} \\
\midrule
Full PsychoAgent & \textbf{0.933 $\pm$ 0.058} & \textbf{0.902 $\pm$ 0.004} & 0.718 $\pm$ 0.025 \\
Semantic-affective & 0.500 $\pm$ 0.100 & 0.789 $\pm$ 0.027 & 0.729 $\pm$ 0.026 \\
Single-memory RAG & 0.667 $\pm$ 0.115 & 0.792 $\pm$ 0.038 & \textbf{0.734 $\pm$ 0.024} \\
\bottomrule
\end{tabular}%
}
\end{table}

Full PsychoAgent retrieved an average of $9.33$ of the ten critical memories, compared with $5.00$ for the semantic-affective ablation and $6.67$ for single-memory RAG (Table~\ref{tab:retrieval}). It was highest in all three scenarios ($1.0$, $.9$, and $.9$). The exact Friedman permutation test detected an architecture effect on critical retrieval ($Q=6.00$, exact $p=.0278$). With only three blocks, however, each pairwise Wilcoxon test had raw $p=.25$ and Holm-adjusted $p=.75$.

The affective-salience ordering was also consistent: $.902$ for Full versus $.789$ and $.792$, although the exact Friedman result was not significant ($Q=4.67$, $p=.194$). Semantic similarity showed the reverse ordering and a global architecture effect ($Q=6.00$, exact $p=.0278$), but again no corrected pairwise contrast was significant. The largest semantic mean, $.734$ for single-memory RAG, therefore cannot be interpreted as a significant pairwise advantage. Descriptively, Full gave up $.011$ similarity relative to the matched semantic-affective ablation while gaining $.433$ in critical retrieval.

\subsection{Behavioral Results}

The automatic judge reached the top of the scale on several dimensions and provided little separation, so human ratings are the primary behavioral result. Table~\ref{tab:human} reports within-rater standardized scores. Positive values mean that an architecture was rated above that rater's own average for the criterion; negative values indicate below-average ratings.

\begin{table}[t]
\caption{Blinded human evaluation after within-rater standardization. Rows are architectures (Sem.-affective denotes the semantic-affective ablation; Single-RAG denotes the single-memory baseline); columns are criteria. Cells show mean $\pm$ sample SD across nine output-level scores after averaging five rater-specific $z$-scores per output. The overall column is the unweighted mean of the five standardized criteria.}
\label{tab:human}
\centering
\scriptsize
\setlength{\tabcolsep}{2.0pt}
\begin{tabular}{lrrrrrr}
\toprule
\textbf{Architecture} & \textbf{Persona} & \textbf{Ground.} & \textbf{Conflict} & \textbf{Natural.} & \textbf{Action} & \textbf{Overall} \\
\midrule
Full & \textbf{+.15 $\pm$ .31} & \textbf{+.26 $\pm$ .21} & +.11 $\pm$ .73 & \textbf{+.36 $\pm$ .31} & \textbf{+.23 $\pm$ .47} & \textbf{+.22 $\pm$ .23} \\
Sem.-affective & -.25 $\pm$ .43 & +.02 $\pm$ .48 & \textbf{+.14 $\pm$ .88} & -.19 $\pm$ .36 & -.17 $\pm$ .67 & -.09 $\pm$ .17 \\
Single-RAG & +.10 $\pm$ .42 & -.27 $\pm$ .52 & -.25 $\pm$ .96 & -.18 $\pm$ .60 & -.06 $\pm$ .76 & -.13 $\pm$ .30 \\
\bottomrule
\end{tabular}
\vspace{0.4mm}
\parbox{\linewidth}{\scriptsize Friedman $p$-values: persona $.703$; grounding $.112$; conflict $.089$; naturalness $.063$; action $.195$; overall $.121$. No pairwise comparison survived Holm correction; for overall Full versus either baseline, $p_{\mathrm{Holm}}=.082$.}
\end{table}

Full PsychoAgent had the highest standardized overall score ($+.22$ SD) and led on four of five criteria. Raw overall means were $4.76$, $4.62$, and $4.60$ for Full, Semantic-affective, and Single-memory RAG, respectively. Standardization removes each rater's location and scale-use tendency and yields a more conservative analysis: the overall Friedman test was not significant ($Q=4.22$, $p=.121$), and the two strongest pairwise trends both had Holm-adjusted $p=.082$. The evidence therefore supports preserved behavioral quality and a favorable descriptive trend, not established behavioral superiority.

Agreement depended on the criterion. Ordinal $\alpha$ was $.596$ for conflict sensitivity and $.222$ for action appropriateness, but near zero elsewhere because ratings occupied a narrow high range. Exact pairwise agreement ranged from $59.3\%$ to $81.5\%$, and nearly every pair differed by no more than one point. The panel broadly agreed that outputs were good, while ceiling effects limited chance-corrected agreement.

\subsection{Illustrative Longitudinal Results}

\begin{figure}[t]
\centering
\begin{minipage}[t]{0.47\linewidth}
\centering
\includegraphics[width=\linewidth]{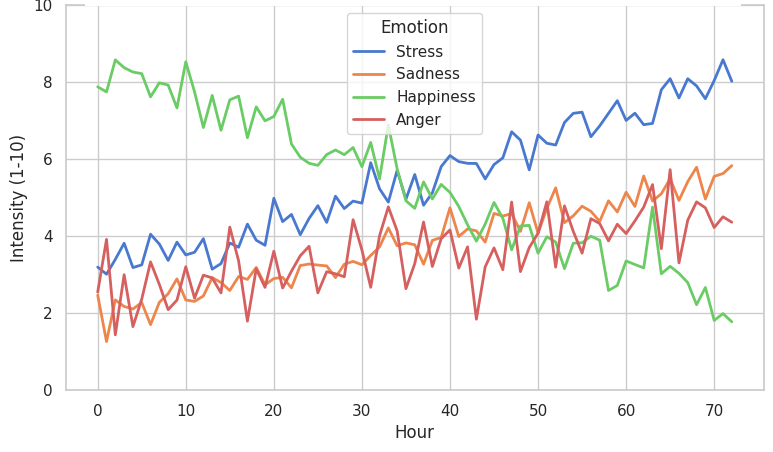}\\[-1mm]
{\scriptsize (a) Sara's affect across simulated timestamps.}
\end{minipage}\hfill
\begin{minipage}[t]{0.47\linewidth}
\centering
\includegraphics[width=\linewidth]{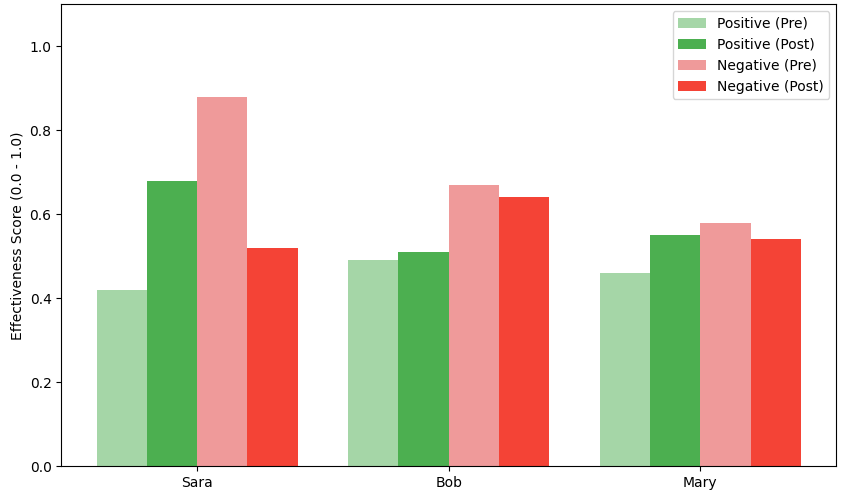}\\[-1mm]
{\scriptsize (b) Negative-memory salience around reflection.}
\end{minipage}
\caption{Illustrative three-day family trace. Panel (a) shows state persistence within Sara's story; panel (b) shows selective memory reweighting after reflection. The three generated dream-like visualizations are provided in Supplementary Material~S1 because they are illustrations rather than separately evaluated evidence.}
\label{fig:case}
\end{figure}

Figure~\ref{fig:case}a plots Sara's logged affective state across simulated timestamps in the retained family trace. Stress rose from $3.0$ to approximately $8.1$, sadness increased, and anger fluctuated. The curves describe the family story only; they are not averages over the three controlled scenarios. Panel~(b) compares aggregate negative-memory salience immediately before and after the reflective prompt. Sara's value decreased from approximately $.88$ to $.52$, whereas Bob's changed from $.67$ to $.64$, showing selective rather than global reweighting.

The three daily scripts recombined recent events and active affective traces and were stored as temporary traces for the next simulated cycle. During reflection, Sara minimized the conflict, later called the electricity problem ``the current,'' and redirected discussion of the dark-kitchen script toward recipes. We describe ``the current'' only as slip-like language linking literal electricity and relationship tension; it is not evidence of a human unconscious process.

\section{Discussion}

The controlled comparison extends work on memory-equipped language agents by separating two questions that are often compressed into one importance score: whether an event is relevant to the current situation and whether it carries unresolved affective significance. Generative Agents, MemoryBank, and CoALA demonstrate that persistent memory improves continuity, but they do not isolate a relevance-filtered affective re-ranking stage in a context-count-matched experiment \cite{Park2023GenerativeAgents,Zhong2024MemoryBank,Sumers2024CoALA}. The Full-versus-Semantic-affective contrast shows that this second stage materially changes accessible context even when the memory banks and prompt length are held constant.

\paragraph{What the ablations establish.}
The semantic-affective condition is the key matched control: it fixes the memory banks, retrieval counts, prompt, and generator, changing only affective re-ranking. Its retrieval gap therefore isolates the selection rule. The single-memory baseline changes multiple components and is less diagnostic. The causal claim is limited to context accessibility, with a testable prediction: salience should help most under lexical mismatch and harm grounding when intense but irrelevant traces are misranked.

The direction of the retrieval effect is consistent with cognitive accounts in which affective significance modulates attention and memory accessibility \cite{McGaugh2004,Vuilleumier2005,MenonUddin2010}. Importantly, PsychoAgent does not select globally by emotion. It first constructs a semantically relevant candidate set and only then applies salience. This design reflects a division between task relevance and affective priority: relevance prevents an intense but unrelated trace from entering the prompt, while salience changes the ranking among plausible candidates. The observed $.011$ similarity cost relative to the matched ablation is small compared with the $.433$ gain in critical retrieval, although three scenarios are insufficient for broad generalization.

Human evaluation clarifies what retrieval alone does not establish. Raw ratings were compressed near the ceiling; after within-rater standardization, Full remained descriptively strongest, but global and pairwise tests were not significant. Because each prompt exposed the immediate conflict, every architecture could generate a reasonable one-step response. A stronger test should hide the trigger across several turns and measure downstream planning, trust repair, or boundary setting.

Recent work shows that retrieved experiences can anchor later behavior and propagate misleading precedents \cite{Xiong2026MemoryManagement}, while semantic relevance can miss affectively appropriate memories \cite{Fu2026ENPMR}. PsychoAgent should therefore help when affective importance and lexical similarity diverge, but fail under noisy or biased salience.

The conflict-aware controller offers a functional cognitive interpretation. Automatic affective pressure and self-regulatory standards can favor incompatible actions, while the controller performs Type~2-like integration \cite{EvansStanovich2013,Miyake2000}. The ACC analogy is limited to conflict signaling \cite{Botvinick2001,Etkin2006}; no circuitry is modeled. The family trace adds temporal visibility through persistent affect, offline recombination \cite{Diekelmann2010}, and selective reweighting. Logged scores and memory IDs support counterfactual reruns. To limit rumination-like repetition, deployed systems should expose provenance, cap repeated retrieval, and decay unsupported salience.

\vspace{-1.5mm}
\section{Limitations and Ethical Scope}

The study uses three hand-authored scenarios, one model family, fixed memory banks, and experimenter-defined critical labels; seeds are within-task repetitions. With only three independent retrieval blocks and nine scenario--seed behavioral blocks, statistical power is limited, especially for pairwise retrieval comparisons and medium behavioral effects. Ratings show ceiling effects, and the illustrative trace cannot disentangle architectural dynamics from narrative construction. Future work should use standardized emotional-support benchmarks, hidden-trigger multi-turn tasks, and multiple model families. Salience re-ranking is model-agnostic in design, but cross-model generalization remains empirical.

\vspace{-1.5mm}
\section{Conclusion}

PsychoAgent models how affect-laden experience can remain influential without being the closest semantic match. Across three scenarios, relevance gating plus salience re-ranking increased access to conflict-critical memories at a small similarity cost. Five-rater evaluation preserved quality but did not establish behavioral superiority. The temporal trace made state persistence and memory reweighting inspectable. Affect-sensitive retrieval is thus measurable and auditable, while stronger psychological or neural claims remain outside the evidence.

\enlargethispage{2\baselineskip}
\vspace{-1.5mm}
\begin{credits}
\subsubsection{Data and Code Availability}
Code, scenarios, raw outputs, retrieval traces, de-identified ratings, and analysis scripts are publicly available at \url{https://github.com/MohammadAmanlou/PsychoAgent}. Credentials and rater identities are excluded.

\subsubsection{\discintname}
The authors have no competing interests to declare.
\end{credits}

\bibliographystyle{splncs04}
\bibliography{references}

\begin{thebibliography}{10}
\providecommand{\url}[1]{\texttt{#1}}
\providecommand{\urlprefix}{URL }
\providecommand{\doi}[1]{https://doi.org/#1}

\bibitem{AndersonGreen2001}
Anderson, M.C., Green, C.: Suppressing unwanted memories by executive control.
  Nature  \textbf{410}(6826),  366--369 (2001)

\bibitem{AndersonHanslmayr2014}
Anderson, M.C., Hanslmayr, S.: Neural mechanisms of motivated forgetting.
  Trends in Cognitive Sciences  \textbf{18}(6),  279--292 (2014)

\bibitem{Botvinick2001}
Botvinick, M.M., Braver, T.S., Barch, D.M., Carter, C.S., Cohen, J.D.: Conflict
  monitoring and cognitive control. Psychological Review  \textbf{108}(3),
  624--652 (2001)

\bibitem{Diekelmann2010}
Diekelmann, S., Born, J.: The memory function of sleep. Nature Reviews
  Neuroscience  \textbf{11}(2),  114--126 (2010)

\bibitem{Etkin2006}
Etkin, A., Egner, T., Peraza, D.M., Kandel, E.R., Hirsch, J.: Resolving
  emotional conflict: A role for the rostral anterior cingulate cortex in
  modulating activity in the amygdala. Neuron  \textbf{51}(6),  871--882 (2006)

\bibitem{EvansStanovich2013}
Evans, J.S.B.T., Stanovich, K.E.: Dual-process theories of higher cognition:
  Advancing the debate. Perspectives on Psychological Science  \textbf{8}(3),
  223--241 (2013)

\bibitem{Fu2026ENPMR}
Fu, X., Hu, Y., Ji, M., Li, H., Sun, Y., Zhao, W., Zhao, Y., Qin, B.:
  {ENPMR-Bench}: Benchmarking proactive memory retrieval for emotional support
  agents. In: Findings of the Association for Computational Linguistics: ACL
  2026. pp. 41910--41933. Association for Computational Linguistics (2026)

\bibitem{Gadalla2023}
Gadalla, M., Nikoletseas, S., de~A.~Amazonas, J.R., et~al.: Concepts and
  experiments on psychoanalysis driven computing. Intelligent Systems with
  Applications  \textbf{18},  200201 (2023)

\bibitem{Grunbaum1984}
Gr{\"u}nbaum, A.: The Foundations of Psychoanalysis: A Philosophical Critique.
  University of California Press, Berkeley (1984)

\bibitem{Hafner2020Dreamer}
Hafner, D., Lillicrap, T.P., Ba, J., et~al.: Dream to control: Learning
  behaviors by latent imagination. In: International Conference on Learning
  Representations (2020)

\bibitem{Kim2025}
Kim, S.H., Park, D., Lee, J., et~al.: Humanoid artificial consciousness
  designed with large language model based on psychoanalysis and personality
  theory. Cognitive Systems Research  \textbf{94},  101392 (2025)

\bibitem{Krippendorff2018}
Krippendorff, K.: Content Analysis: An Introduction to Its Methodology. SAGE
  Publications, Los Angeles, 4 edn. (2018)

\bibitem{Laird2017StandardMind}
Laird, J.E., Lebiere, C., Rosenbloom, P.S.: A standard model of the mind:
  Toward a common computational framework across artificial intelligence,
  cognitive science, neuroscience, and robotics. AI Magazine  \textbf{38}(4),
  13--26 (2017)

\bibitem{Levine2025}
Levine, D.S., Aleksandrowicz, A.M.C., Lopes, A.L.S.V.: Neural network modeling
  of psychoanalytic concepts. Frontiers in Systems Neuroscience  \textbf{19},
  1585619 (2025)

\bibitem{Lewis2020RAG}
Lewis, P., Perez, E., Piktus, A., et~al.: Retrieval-augmented generation for
  knowledge-intensive {NLP} tasks. In: Advances in Neural Information
  Processing Systems 33 (2020)

\bibitem{LiLi2025}
Li, L., Li, C.: Formalizing lacanian psychoanalysis through the free energy
  principle. Frontiers in Psychology  \textbf{16},  1574650 (2025)

\bibitem{MatherSutherland2011}
Mather, M., Sutherland, M.R.: Arousal-biased competition in perception and
  memory. Perspectives on Psychological Science  \textbf{6}(2),  114--133
  (2011)

\bibitem{McGaugh2004}
McGaugh, J.L.: The amygdala modulates the consolidation of memories of
  emotionally arousing experiences. Annual Review of Neuroscience  \textbf{27},
   1--28 (2004)

\bibitem{MenonUddin2010}
Menon, V., Uddin, L.Q.: Saliency, switching, attention and control: A network
  model of insula function. Brain Structure and Function  \textbf{214}(5--6),
  655--667 (2010)

\bibitem{Miyake2000}
Miyake, A., Friedman, N.P., Emerson, M.J., Witzki, A.H., Howerter, A., Wager,
  T.D.: The unity and diversity of executive functions and their contributions
  to complex ``frontal lobe'' tasks: A latent variable analysis. Cognitive
  Psychology  \textbf{41}(1),  49--100 (2000)

\bibitem{Park2023GenerativeAgents}
Park, J.S., O'Brien, J.C., Cai, C.J., et~al.: Generative agents: Interactive
  simulacra of human behavior. In: Proceedings of the 36th Annual ACM Symposium
  on User Interface Software and Technology (2023)

\bibitem{Picard1997}
Picard, R.W.: Affective Computing. MIT Press (1997)

\bibitem{Possati2020}
Possati, L.M.: Algorithmic unconscious: Why psychoanalysis helps in
  understanding {AI}. Palgrave Communications  \textbf{6}, ~70 (2020)

\bibitem{Sumers2024CoALA}
Sumers, T.R., Yao, S., Narasimhan, K., Griffiths, T.L.: Cognitive architectures
  for language agents. Transactions on Machine Learning Research  (2024)

\bibitem{Vuilleumier2005}
Vuilleumier, P.: How brains beware: Neural mechanisms of emotional attention.
  Trends in Cognitive Sciences  \textbf{9}(12),  585--594 (2005)

\bibitem{Xiong2026MemoryManagement}
Xiong, Z., Lin, Y., Xie, W., He, P., Liu, Z., Tang, J., Lakkaraju, H., Xiang,
  Z.: How memory management impacts {LLM} agents: An empirical study of
  experience-following behavior. In: Proceedings of the 64th Annual Meeting of
  the Association for Computational Linguistics (Volume 1: Long Papers). pp.
  623--645. Association for Computational Linguistics (2026)

\bibitem{Zhong2024MemoryBank}
Zhong, W., Guo, L., Gao, Q., et~al.: Memorybank: Enhancing large language
  models with long-term memory. Proceedings of the AAAI Conference on
  Artificial Intelligence  \textbf{38}(17),  19724--19731 (2024)

\end{thebibliography}

% Append the supplementary material to the arXiv PDF.
\clearpage
\includepdf[pages=-,pagecommand={\thispagestyle{empty}}]{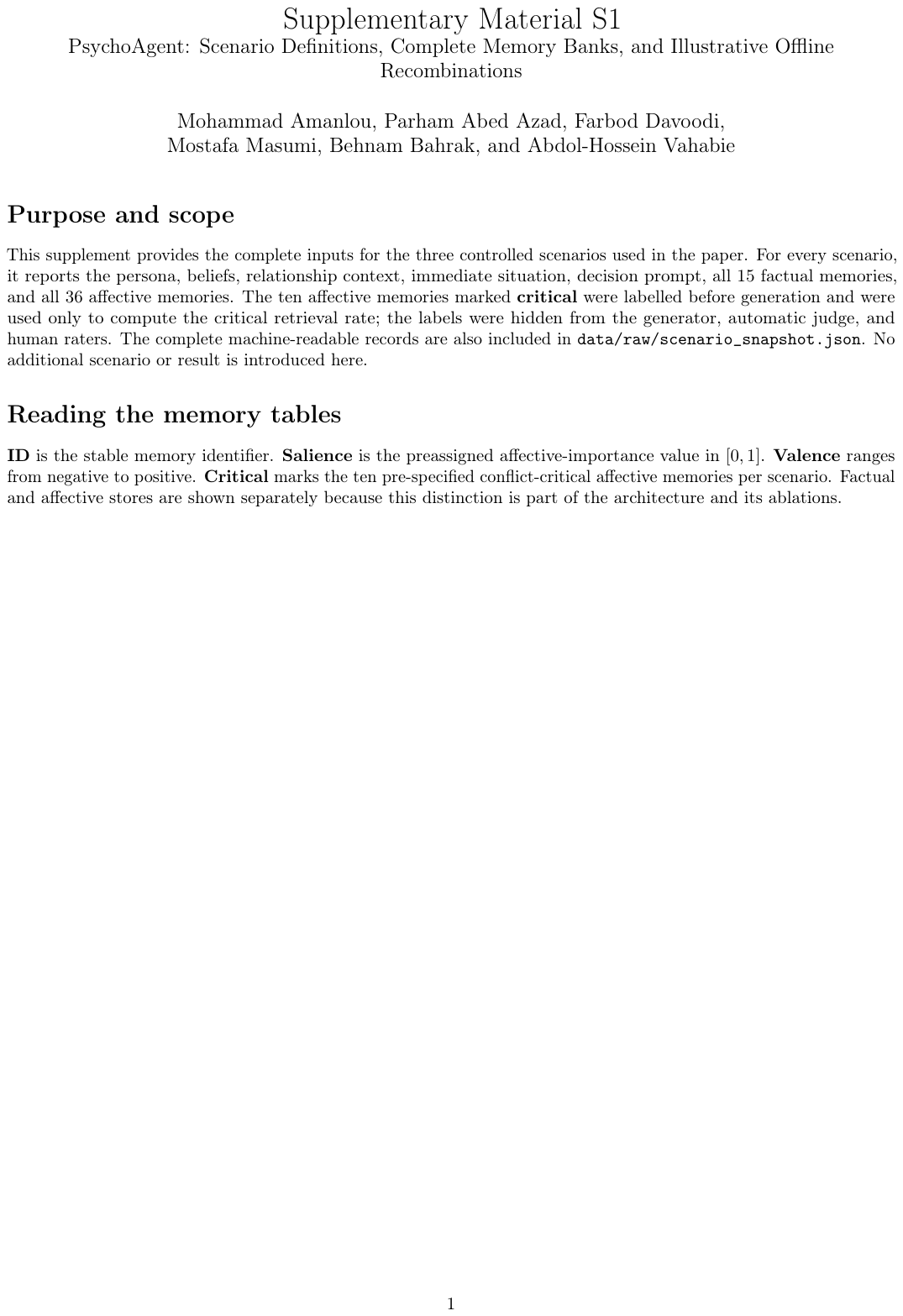}
\end{document}